\documentclass[reprint,showpacs,amsmath,superscriptaddress,amssymb,aps,prx,floatfix,longbibliography,nofootinbib]{revtex4-2}
\usepackage{amsthm}
\usepackage{amsmath, mathtools}
\usepackage[utf8]{inputenc}	
\usepackage[english]{babel}
\usepackage{amsmath}
\usepackage{graphicx}		
\usepackage{natbib}
\usepackage{textcomp}
\usepackage{gensymb}
\usepackage[usenames,dvipsnames,svgnames,table]{xcolor}
\usepackage[hidelinks,colorlinks=false,urlcolor=Cerulean,citecolor=black]{hyperref}
\usepackage{siunitx}
\usepackage{mathrsfs}
\usepackage{multirow}
\usepackage{bm}
\usepackage{xfrac}
\usepackage{comment}
\usepackage[normalem]{ulem}

\renewcommand{\i}{\mathrm{i}}

\DeclareMathOperator{\C}{\mathbb{C}}

\newcommand{\revision}[1]{\textcolor{black}{#1}}

\begin{document}

\title{An exact mathematical description of computation with transient \\ spatiotemporal dynamics \revision{in a complex-valued neural network}}

\author{Roberto C. Budzinski}
\thanks{These authors contributed equally}
\affiliation{Department of Mathematics, Western University, London, ON, Canada}
\affiliation{Western Institute for Neuroscience, Western University, London, ON, Canada}
\affiliation{Western Academy for Advanced Research, Western University, London, ON, Canada}
\author{Alexandra N. Busch}
\thanks{These authors contributed equally}
\affiliation{Department of Mathematics, Western University, London, ON, Canada}
\affiliation{Western Institute for Neuroscience, Western University, London, ON, Canada}
\affiliation{Western Academy for Advanced Research, Western University, London, ON, Canada}
\author{Samuel Mestern}
\affiliation{Graduate Program in Neuroscience, Western University, London, Canada}
\author{Erwan Martin}
\affiliation{Department of Mathematics, Western University, London, ON, Canada}
\affiliation{Western Institute for Neuroscience, Western University, London, ON, Canada}
\affiliation{Western Academy for Advanced Research, Western University, London, ON, Canada}
\author{Luisa H. B. Liboni}
\affiliation{Department of Mathematics, Western University, London, ON, Canada}
\affiliation{Western Institute for Neuroscience, Western University, London, ON, Canada}
\affiliation{Western Academy for Advanced Research, Western University, London, ON, Canada}
\author{Federico W. Pasini}
\affiliation{Huron University College, London, ON, Canada}
\author{J\'an Min\'a{\v c}}
\affiliation{Department of Mathematics, Western University, London, ON, Canada}
\affiliation{Western Academy for Advanced Research, Western University, London, ON, Canada}
\author{Todd Coleman}
\affiliation{Department of Bioengineering, Stanford University, Stanford, CA, USA}
\author{Wataru Inoue}
\affiliation{Robarts Research Institute, Western University, London, Canada}
\affiliation{Department of Physiology and Pharmacology, Western University, London, Canada}
\author{Lyle E. Muller}
\email{lmuller2@uwo.ca}
\affiliation{Department of Mathematics, Western University, London, ON, Canada}
\affiliation{Western Institute for Neuroscience, Western University, London, ON, Canada}
\affiliation{Western Academy for Advanced Research, Western University, London, ON, Canada}

\begin{abstract}
We study a complex-valued neural network (cv-NN) with linear, time-delayed interactions. We report the cv-NN displays sophisticated spatiotemporal dynamics, including partially synchronized ``chimera'' states. We then use these spatiotemporal dynamics, in combination with a nonlinear readout, for computation. The cv-NN can instantiate dynamics-based logic gates, encode short-term memories, and mediate secure message passing through a combination of interactions and time delays. The computations in this system can be fully described in an exact, closed-form mathematical expression. Finally, using direct intracellular recordings of neurons in slices from neocortex, we demonstrate that \revision{computations} in the cv-NN are decodable by living biological neurons. These results demonstrate that complex-valued linear systems \revision{can perform sophisticated computations, while also being exactly solvable. Taken together, these results open future avenues for design of highly adaptable, bio-hybrid} computing systems that can interface seamlessly with other neural networks.
\end{abstract} 

\maketitle

\section{Introduction}

Spatially extended dynamics represent a powerful substrate for computation. Neural systems perform sensory computations with organized spatiotemporal dynamics traveling over maps of sensory space \cite{ermentrout2001traveling,muller2018cortical,benigno2023waves}. For example, traveling waves of spontaneous neural activity traverse highly organized cortical maps of visual space, modulating perceptual sensitivity as they travel across local neural circuits \cite{davis2020spontaneous}. In non-biological systems, spatiotemporal patterns of optical or electromagnetic waves can perform sophisticated computations, such as predicting input sequences \cite{vandoorne2014experimental} or performing transformations \cite{del2018leveraging}. Despite their many differences, these example systems all perform computations through sophisticated spatiotemporal dynamics. In general, nonlinearities in these systems are thought to be essential for spatiotemporal computation, because they can provide a rich diversity of dynamical behavior that can, in turn, be leveraged for computation \cite{kia2017nonlinear,zanin2011computation}. Nonlinear dynamics are also used extensively in machine learning for training neural networks to perform specific tasks \cite{yang2019task,kim2023neural} and in physics for training reservoir computers to predict chaotic dynamics \cite{pathak2018model,tanaka2019recent}. \revision{To understand how these systems learn, analytical techniques such as mean-field approaches have created important insights into macroscopic features of the dynamics in these nonlinear networks \cite{helias2020statistical,keup2021transient}, such as their autocorrelation structure and transitions to chaos \cite{sompolinsky1988chaos,bordelon2022population,keup2021transient,aljadeff2015transition}. Despite these advances, however, it remains difficult to understand the precise dynamical trajectory an individual nonlinear system uses to perform a specific task. For example, a trained recurrent neural network has a set of connection weights that can perform a perceptual decision-making task \cite{song2016training}, where the network can integrate inputs over a long time before converging to a choice, but the precise dynamical trajectory by which the network completes the task remains difficult to access mathematically. This is due to the fact that} the nonlinear systems that are useful for computation do not have closed-form mathematical solutions, which could provide fundamental insight into how these networks perform tasks and eliminate the need for training paradigms that can consume large amounts of energy. For these reasons, a major recent focus has been to implement spatiotemporal computation in real-world physical systems \cite{rafayelyan2020large}, which could substantially speed up network training and reduce energy consumption. While these physical systems have provided significant advances over neural networks trained on a digital computer, they still require substantial effort to design because they are based on sophisticated nonlinear dynamics. The key missing element is a system that is sophisticated enough to perform computations, while also providing mathematical insight. While linear systems allow for closed-form mathematical expressions, they are often thought to be too simple to produce sophisticated spatiotemporal dynamics \cite{jiang2019irrelevance,kia2017nonlinear}. \revision{Nonlinear systems have, in turn, more often been the focus for computation.}

\revision{Here, we study a system with linear dynamics and nonlinear readout. Several experimental and numerical simulation studies have observed that systems with this structure can be useful for computation \cite{vandoorne2014experimental,vinckier2015high,laporte2018numerical,lugnan2020photonic,ma2021addressing}. We report that the full computation in our system can be solved exactly. The exact solution allows designing specific computations in the system with a precise mathematical formulation, and further, provides fundamental theoretical insight into computation with linear dynamics and nonlinear readout. Specifically, time delays in the network interactions extend the window of time for which amplitudes in this network remain bounded, and during this window, the network displays rich spatiotemporal dynamics. It is these rich spatiotemporal dynamics that we find can be used for computation. These} dynamics include the well-studied ``chimera'' states, in which clusters of order and disorder can emerge in networks where all nodes are connected in exactly the same way \cite{abrams2004chimera,motter2010spontaneous}. It is surprising to find chimera states in a linear system, because they are usually studied in nonlinear oscillator networks \cite{omelchenko2013nonlocal,kotwal2017connecting,panaggio2015chimera} or reaction-diffusion systems \cite{tinsley2012chimera,totz2018spiral}. That they can occur in a linear system opens a novel avenue for studying computation with transient dynamics, which are known to be important for neural computation \cite{maass2002real,jaeger2004harnessing,rabinovich2008transient}. Here, we describe, with a precise mathematical expression, the process of computation with transient network states, for the first time.

The dynamics of this complex-valued neural network (cv-NN) is governed by the  differential equation
\begin{equation}
    \dot{\bm{x}}(t) = \big(\i \omega \bm{I} + \bm{K} \big) \bm{x}(t),
    \label{eq:differential_eq}
\end{equation}
where $\bm{x}(t) \in \C^{N}$ is a complex vector that specifies the state of network at each point in time, \revision{and} $N$ is the number of elements in the cv-NN. \revision{Though in this work we focus on the general computational properties of this system, each node in the network can be thought to represent the activation of a small patch of neurons in a single region, as in previous work on complex-valued models to approximate the dynamics of biological spiking networks \cite{schaffer2013complex}.} $\bm{I}$ is the identity matrix, $\i$ is the imaginary unit\footnote{Note the difference between the imaginary unit $\i$ and the index variable $i$.}, \revision{and} $\omega$ specifies the frequency at which the nodes' dynamics evolve. Throughout this work, we set $\omega = 2\pi f$ to have a natural frequency of $f = 10\,$Hz. This frequency sets a timescale relevant to information processing in the brain. Other values of $f$ can be chosen without loss of generality. We have previously shown that this complex-valued system displays the hallmarks of canonical synchronization behavior found in oscillator networks \cite{budzinski2022geometry,budzinski2023theory}, and while we consider the case of homogeneous natural frequencies here, the approach also generalizes to heterogeneous natural frequencies \cite{muller2021algebraic, budzinski2023analytical}.

It is well established that a fixed time delay can be approximated by a phase delay in the oscillator interaction term \cite{jeong2002time,ko2007effects}. The matrix $\bm{K} = \epsilon e^{-\i \phi} \bm{A}$ collects information on interactions between the nodes, where $\epsilon$ is the coupling strength and $\phi = \omega T$ is a phase delay approximating time delay $T$. The matrix $\bm{A}$ represents connection weights $a_{ij}$ between nodes $i$ and $j$ in the network  (see Appendix A). Here, we consider the nodes in the cv-NN to be coupled in a one-dimensional ring with periodic boundary conditions where the connection weight between two nodes decays in a distance-dependent fashion. To implement dynamics based computation, we set up a system with an input layer, the cv-NN, and a decoder that interprets the phase dynamics of the network (Fig.\,\ref{fig:example_computer}a). \revision{We utilize polar notation for complex numbers throughout the text, which provides a direct way to analyze the phase dynamics that will be used for computation. Specifically, the cv-NN performs computations with spatiotemporal dynamics in the recurrent network, in combination with a nonlinearity in the readout.} With this formulation, we can now write the entire computation in the system with the following closed-form expression:
\begin{equation}
    \label{eq:computation}
    o_k(t) = \Theta_\sigma R_k \mathcal{D}_t \bm{x}(0),
\end{equation}
where the operator $\mathcal{D}_t = e^{i\omega t}e^{\bm{K}t}$ represents an exact solution for the linear dynamics in Eq.\,(\ref{eq:differential_eq}), starting from initial conditions $\bm{x}(0)$ (see Appendix B), the operator $R_k:\mathbb{C}^N \rightarrow \mathbb{R}$ quantifies the level of synchronization in a local patch $k$ of the network (see Appendix C, Eq.\,(A4)), and $\Theta_\sigma$ is the standard Heaviside function shifted by a threshold $\sigma$. Here, $o_k(t)$ represents the output of decoder node $k$. This equation captures the computation in the cv-NN in a closed-form analytical expression, in terms of linear dynamics and nonlinear readout working on the network's initial state $\bm{x}(0)$.

\section{Results}

\begin{figure}[t!]
    \centering
    \includegraphics[width=0.85\columnwidth]{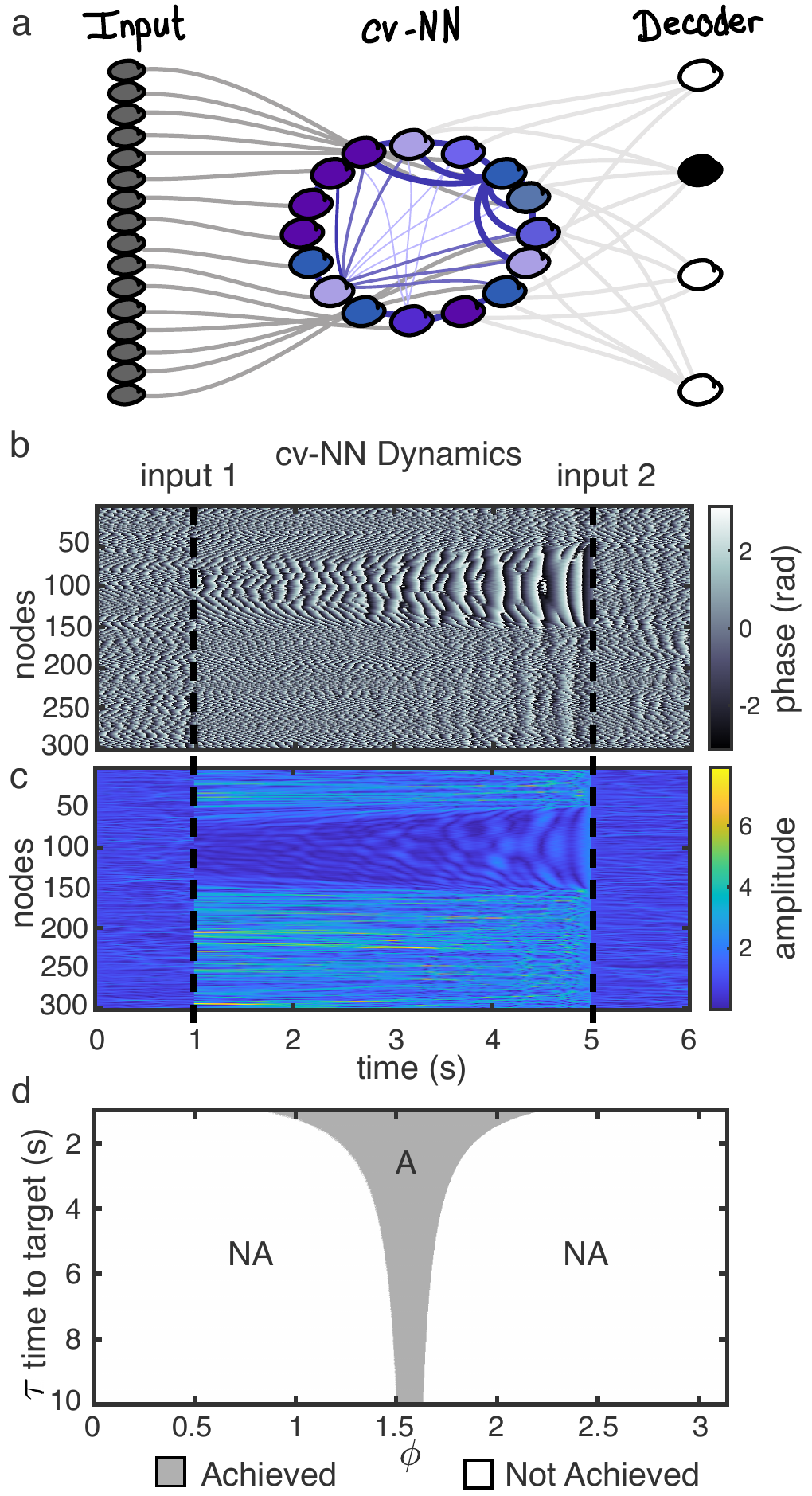}
    \caption{\textbf{A cv-NN with linear dynamics exhibits sophisticated spatiotemporal patterns.} \textbf{(a)} Our framework is composed of an input layer connected to a recurrent layer (the cv-NN) that generates dynamics which are then interpreted by a decoder. \textbf{(b-c)} We use the inverse of operator $\mathcal{D}$ to determine the input required to drive the network to a specific target pattern at a precise time, several seconds into the future (see Appendix B). The dynamics of the cv-NN starts in an asynchronous state due to random initial conditions. After input 1, the network evolves to a chimera state pattern. A second input leads the cv-NN back to an asynchronous state. The cv-NN exhibits both \textbf{(b)} phase and \textbf{(c)} amplitude dynamics. Throughout this work, we use the phase dynamics to perform computations. \textbf{(d)} The success of this process -- whether or not the cv-NN achieves the target pattern at the desired time -- is quantified by the similarity metric Eq. (\ref{eq:similarity}). There is a specific range of the delay parameter, $\phi$, and $\tau$ for which the target pattern is achieved by the cv-NN dynamics.}
    \label{fig:example_computer}
\end{figure}
To describe the system in more detail, we consider computation in the cv-NN in an input-decoder framework (Fig.\,\ref{fig:example_computer}a), where the $N$ nodes in the network receive connections from $M$ input nodes, and project to a set of $L$ output nodes that constitute a decoder. The set of weighted connections from the input nodes to the network are collected into an $M \times N$ weight matrix $\mathbf{W}$, which can take a variety of forms. In one setup, each input node could project to a single network node. In this case, $\mathbf{W}$ \revision{is the $N \times N$ identity matrix $\mathbf{I}_N$,} and all input nodes drive the network with specific complex numbers. In another setup, each input node may project to the full network with varying weights. In this case, $\mathbf{W}$ is in the set of $M \times N$ matrices with complex coefficients ($\mathbf{W} \in \mathcal{M}_{M\times N}(\C)$), and input nodes are either ``on" or ``off". In either case, the input to the network takes the form of a complex valued vector, which is applied through multiplication with the state vector of the cv-NN (see Fig.\,A2). This process can be thought of as ``nudging” the state of the network onto the trajectory in state space that will allow it to evolve with the desired dynamics. Since the dynamics of the cv-NN is governed by the operator $\mathcal{D}$, the input weights required for the system to evolve to a specific target pattern can be computed precisely using the inverse operator $\mathcal{D}^{-1}$\,(see Appendix B and supplement Fig. S1). Starting from random and asynchronous initial conditions, we can calculate the input needed for the system to evolve to an arbitrary state several seconds into the future (``input 1'', Figs.\,\ref{fig:example_computer}b, \ref{fig:example_computer}c). Adding this input to the state vector of the cv-NN brings the network to the correct state required for evolution to the target. We can then use these inputs to design specific dynamics in the cv-NN, in both amplitude and phase, and we use the phase dynamics for computation throughout this paper.

Figure \ref{fig:example_computer} illustrates an example of this input-output framework in the cv-NN, depicting both the phase (Fig.\,\ref{fig:example_computer}b) and amplitude (Fig.\,\ref{fig:example_computer}c) at each node in the network. In this example, input 1 drives the network to a partially phase synchronized state 4 seconds in the future (Fig.\,\ref{fig:example_computer}b, Movie S1). This partially synchronized state, where a specific group of nodes has the same phase while other nodes remain in an asynchronous state, represents a chimera state \cite{abrams2004chimera}. Chimera states have been studied in many nonlinear systems -- e.g.\,reaction-diffusion systems \cite{tinsley2012chimera,totz2018spiral}, recurrent neural networks \cite{masoliver2022embedded}, and networks of Kuramoto oscillators \cite{abrams2004chimera, panaggio2015chimera, kotwal2017connecting}. \revision{We report that the linear cv-NN displays a range of chimera states, from the short-lived states that we use here for computation (Fig.\,\ref{fig:example_computer}b) to chimeras that exist for long timescales (Fig.\,S2).} Because we now have a system that exhibits sophisticated dynamics \revision{such as these transient chimeras}, and has a closed-form expression, we can study the input-output mappings in the cv-NN and design dynamics to perform computation.

Not all desired target states are achievable with the network dynamics, however. For instance, if the target state is a chimera, but the network is in a fast-synchronizing regime, the network dynamics will not match the target state. With this in mind, we introduce a similarity measurement: 
\begin{equation}
    \mathcal{S} = \frac{1}{N}\Bigg|\sum_{j=1}^{N} e^{\i \mathrm{Arg}[{\chi_{j}]} } e^{-\i \mathrm{Arg}[{x}_{j}(\tau)]}\Bigg|
    \label{eq:similarity}
\end{equation}
to quantify the match between the target phase pattern $\mathrm{Arg}[\bm{\chi}]$, designed to appear at time $\tau$, and the phase pattern displayed by the system at that time, $\mathrm{Arg}[\bm{x}(\tau)]$. In general, amplitudes in the linear cv-NN will grow to become unbounded or decay to zero during a transient time window. Further, in many configurations the network will quickly synchronize, which collapses all inputs to the same output state (see Appendix D, Fig.\,A4). We find, however, that for intermediate interaction strength $\epsilon$ and values of $\phi$ within an interval near $\sfrac{\pi}{2}$, this procedure can generate arbitrary spatiotemporal patterns up to $10$ s into the future (Fig.\,1d) (see also Appendix B, and  Figs. S1 and A5). It is important to note that, while here we consider computations based on phase synchronized clusters in the network, this framework naturally generalizes to different phase patterns.
\begin{figure}[t!]
    \centering
    \includegraphics[width=\columnwidth]{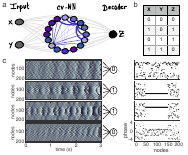}
    \caption{\textbf{The cv-NN can perform simple computations.} \textbf{(a)} Here, we consider two binary inputs ``X" and ``Y" applied to the cv-NN, and we use its dynamics to compute the output ``Z" that is interpreted by the decoder. When one (or both) of the inputs turns on, the input specified by the weighted connections between the input node and the network is added to the state vector of the cv-NN. \textbf{(b)} Here, we implement an XOR gate. Rows in this table represent the input node state vector, $[X,Y]$, and the resulting output, $Z$. \textbf{(c)} when X = 1 and Y = 0, or X = 0 and Y = 1 (i.e. precisely one input is applied), we observe a coherent cluster in the spatiotemporal dynamics. This phase synchronized cluster is recognized by the decoder, which returns Z = 1. However, when X = 0 and Y = 0, or X = 1 and Y = 1 (i.e. neither input is applied or both inputs are applied), no phase synchronized cluster appears, and the decoder returns Z = 0.}
    \label{fig:xor_gate}
\end{figure}

This result demonstrates the cv-NN can achieve a target state at a specific time window in the future. To demonstrate that these states can be used for computation, we implement the cv-NN with two possible inputs, X and Y, and one output, Z, which is decoded from the network phase dynamics (Fig.\,\ref{fig:xor_gate}a, Movie S2). This setup allows for the realization of an XOR logic gate (Fig.\,\ref{fig:xor_gate}b). When X and Y are both $0$, the cv-NN remains asynchronous, and no chimera occurs (Z$=0$) (Fig.\,\ref{fig:xor_gate}c, top). When either input X or Y is $1$, a synchronized cluster occurs in the center of the network (Z$=1$) (Fig.\,\ref{fig:xor_gate}c, middle two rows). Lastly, when X and Y are both $1$, these competing inputs interfere in such a way that no synchronized cluster is observed in the network (Fig.\,\ref{fig:xor_gate}c, bottom). It is important to note that, in contrast to the way computations are often implemented in neural networks, where nonlinearities at single neurons alternate with pooling across trained network connections, the interference underlying the mapping (X=1, Y=1) $\rightarrow$ Z=0 occurs in the linear dynamics of the cv-NN, with the nonlinearity $\Theta_\sigma R_k$ only applied once at the readout. This specific combination of linear dynamics and simple nonlinear readout allows specifying the XOR operation precisely in a closed-form expression (Eq.~(\ref{eq:computation})). This XOR gate is robust to noise (see Appendix F and Supplement Fig.\,S\revision{3}). Having a closed-form expression opens the opportunity to generalize easily to other standard logic gates (Supplement Fig.\,S\revision{4}) and, potentially, more complex logic operations, in a natural way.

\begin{figure}[b!]
    \centering
    \includegraphics[width=.9\columnwidth]{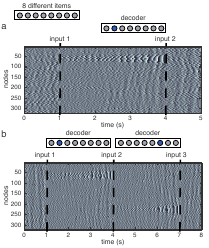}
    \caption{\textbf{The cv-NN can perform short-term memory tasks and online updating.} \textbf{(a)} We use our computational framework to perform a short-term memory task in which $1$ of $8$ possible items is to be held in working memory for $3$ seconds. The remembered item is encoded in the cv-NN dynamics by the location of the phase synchronized cluster, and can be read out by a simple decoder. We obtain analytically the input necessary to drive the cv-NN dynamics to the specific pattern for each of the $8$ items. The input can be understood as a cue in the context of this task. In the example shown here, the input cuing item $2$ is applied, and the cv-NN dynamics depicts the corresponding chimera state. Once the decoder successfully interprets the phase dynamics, a second input (input 2) then drives the network back to asynchronous behavior. \textbf{(b)} This approach also allows memories to be updated online. In this example, due to the first cue, the network initially stores item $2$ in memory. The application of input 2 (second cue) updates this memory to item $6$. After the item is decoded, input 3 drives the cv-NN dynamics back to an asynchronous state. Our approach naturally generalizes to online updating because we do not need any information about the future state of the cv-NN to make updates, since we have a closed-form mathematical expression describing the whole process.}
    \label{fig:short_term_memory}
\end{figure}

\begin{figure*}[t!]
    \centering
    \includegraphics[width=0.95\textwidth]{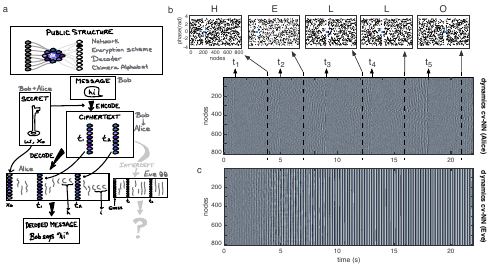}
    \caption{\textbf{The cv-NN enables message transmission based on spatiotemporal dynamics.} \textbf{(a)} By associating letters to specific spatiotemporal patterns (different phase synchronized clusters that form a public alphabet -- see Appendix G), we can use our framework for message transmission. In this example, we consider the traditional scenario where Bob sends a message to Alice, which Eve tries to intercept. Here, the public structure is given by the parameters $\epsilon$, $\alpha$, $\phi$ and the matrix $\bm{A}$. Further, Bob and Alice agree in advance on the parameter $\omega$ and the initial state of the cv-NN $\bm{x}({0})$, which configures the secret key. To encrypt his message, Bob decides on a set of target times when the phase clusters corresponding each letter of his message should appear. He chooses a set of input times $t_{j}$, and uses the inverse operator $\mathcal{D}^{-1}$ to obtain the required inputs $\mathcal{I}_{j}$. He then sends the ciphertext \{$\mathcal{I}_{j}$,$t_{j}$\} to Alice. \textbf{(b)} Alice then implements the cv-NN using the secret key \{$\omega$, $\bm{x}({0})$\} and applies the inputs $\mathcal{I}_{j}$ at the times $t_{j}$. The resulting spatiotemporal dynamics of the cv-NN can then be interpreted by the decoder using the public alphabet. By applying the shared inputs, different phase clusters can be observed in the cv-NN dynamics, which result in the message ``HELLO". \textbf{(c)} At the same time, Eve uses the public information to build the cv-NN and tries to intercept the message. Even though Eve is able to obtain the ciphertext, she is not able to decode the message: when she applies the inputs $\mathcal{I}_{j}$ at the correct times $t_{j}$, she does not obtain the phase clusters in the dynamics of her cv-NN because she does not have access to the secret key $\{\omega, \bm{x}({0})\}$.} 
    \label{fig:message_transmission}
\end{figure*}
The cv-NN can thus perform simple spatiotemporal computations by ``holding'' target states several seconds into the future. These target states could, conceivably, enable a form of in-memory computation, which has proven to be a promising departure from traditional models of von Neumann computing architectures \cite{sebastian2020memory}. To explore the possibility of performing in-memory computations with target states in the cv-NN, we considered an example task in which $1$ of $8$ items is to be held in short-term memory for $3$ seconds. An input to the cv-NN cues the item to be remembered, which is encoded by a specific pattern in the cv-NN dynamics, and can then be read out by decoder units with connections to nearby nodes in the network (Fig. \ref{fig:short_term_memory}a). As before, we store the item in a coherent phase cluster at a specific position in the cv-NN. This cluster then triggers the decoding unit corresponding to the item held in memory (see Appendix C). The network is initially asynchronous, due to random initial conditions ($t < 1$ second, Fig. \ref{fig:short_term_memory}a). Following a specific input that cues item $2$ at $t = 1$ second, the network evolves to a state with a coherent phase cluster centered at decoder $2$, representing the item held in memory. After the item is correctly decoded, another input is applied, and the cv-NN returns to an asynchronous state ($t > 4$ seconds, Fig. \ref{fig:short_term_memory}a). As with the implementation of logic gates, this framework for short-term memory is robust to noise and perturbation (Supplement Fig.\,S4).

A key feature of in-memory computation is the ability to update online, a feature shared with biological working memory \cite{morris1990memory}. For instance, if someone is asked to keep a phone number in memory, it is also possible for them to update the last digit from a ``1'' to a ``9''. Online updates provide biological working memory with the flexibility to adapt to inputs and solve problems over extended time scales \cite{oreilly2006}. To demonstrate online updating, we consider a longer task where the cv-NN must switch between items $2$ and $6$ after $4$ seconds (Fig.\,\ref{fig:short_term_memory}b, Movie S3). Importantly, the input cue needed for the switch is given by a single vector that can be computed locally in time, without requiring future information about the cv-NN's state. These results demonstrate that the cv-NN can store short-term memories with a process that can be both updated online and described with a mathematically exact solution.

Short-term states in the cv-NN can also be used to encode and transmit information between two or more sources, in a simple symmetric-key encryption format \cite{Delfs2015}. To demonstrate this, we mapped different coherent phase clusters to each letter of the English alphabet (and one to a blank space), which leads to $27$ different patterns constituting a ``chimera alphabet'' (see Appendix G). We then consider the traditional scenario in which Bob sends a message to Alice, which Eve tries to intercept (Fig.\,\ref{fig:message_transmission}a). Bob and Alice agree on a secret key, $\{\omega, \bm{x}(0)\}$, where $\omega$ denotes the intrinsic frequency and $\bm{x}(0)$ the initial conditions. The chimera alphabet and the network structure ($\bm{K}$) form the public structure through which the message is transmitted. Eve therefore knows the full transmission framework, including $\bm{K}$, the chimera alphabet, and the format of the ciphertext, but must attempt to guess the secret key shared between Alice and Bob. To encrypt a message, Bob first chooses a set of target times at which the encoded letters should appear in Alice's cv-NN. He then uses $\mathcal{D}^{-1}$ to compute the set of inputs ($\mathcal{I}_j$) to apply at specific input times ($t_j$) so that the chimera letters appear as desired (see Appendix B). Bob sends Alice the ciphertext $\{\mathcal{I}_j,t_j\}$. Alice initializes her cv-NN using the shared secret $\{\omega, \bm{x}(0)\}$, then applies the ciphertext, letting the network dynamics evolve according to the operator $\mathcal{D}$. Because Alice has the secret key, synchronized phase clusters will appear. When this happens, Alice decodes each letter of the message using the public chimera alphabet (Fig. \ref{fig:message_transmission}b, Movie S4). We note that the synchronized clusters appear at the target times chosen by Bob, but that these times do not need to be known by Alice and are discovered in the spatiotemporal dynamics of the cv-NN. At the same time, an eavesdropper, Eve, intercepts the ciphertext and applies the inputs to the public network to decode the encrypted message; however, because Eve does not have the secret key $\{\omega, \bm{x}(0)\}$, Eve's network does not reach the chimera states and will, in practice, evolve to synchronized states.

This example of dynamics-based encryption is robust to random attacks. Randomly guessing $\omega$ and $\bm{x}_{0}$ does not produce phase clusters from the alphabet (Fig.\,A7). Further, the inputs $\mathcal{I}_j$ must be applied in the correct sequence and at the correct times $t_j$; otherwise, the target patterns are not obtained. Finally, one may question whether the synchronization of Eve's network (Fig. \ref{fig:message_transmission}c) could offer her some insight into the private information. This is not the case, however, because in practice Bob and Alice can always extend the private information to a series of keys $\{\omega, \bm{x}(0)\}$ and jump between these keys in a sequence \cite{mitola1999cognitive,wang2010advances}.
\begin{figure}[b!]
    \centering
    \includegraphics[width=0.98\columnwidth]{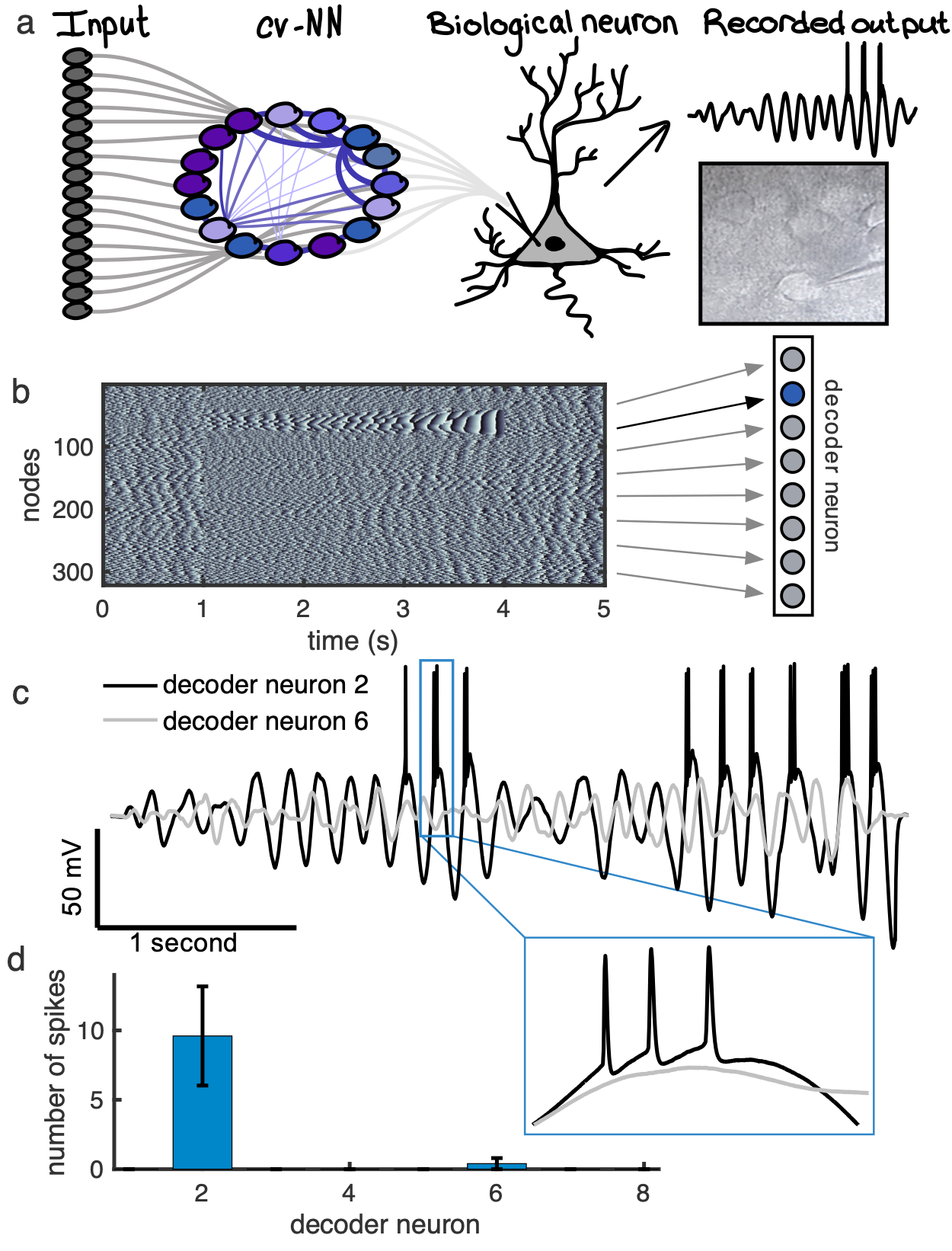}
    \caption{\textbf{Biological neurons can successfully decode the dynamics of the cv-NN.} \textbf{(a)} We implement the cv-NN using real, biological neurons as the decoders. To do so, we inject the dynamics of the cv-NN as a current into a biological neuron, whose resulting physiological signal is used for decoding. \textbf{(b)} As an example, we consider the short-term memory task represented in Fig. \ref{fig:short_term_memory}. The location of the phase synchronized cluster indicates which item is being remembered (item 2 in this case). \textbf{(c)} The current input created from the segment of the network corresponding to the phase synchronized cluster causes the biological neuron decoder to fire repeatedly (black trace). However, when the current input corresponds to an asynchronous segment of the network, the neuron does not fire (gray trace). \textbf{(d)} This procedure is repeated for several different trials, which shows successful decoding.}
    \label{fig:biological_neurons}
\end{figure}

These results demonstrate the cv-NN can perform computations, store short-term memories, and can enable secure message passing, but is it possible to communicate with a biological neuron? We next test whether biological neurons can decode the spatiotemporal dynamics underlying computation in the cv-NN. To do this, we injected the cv-NN dynamics directly as a current into a biological cell via an intracellular recording electrode (Fig.~\ref{fig:biological_neurons}a, see also Appendix H and Fig.\,A8), and then used the resulting spikes generated by the cell as the decoder (see Appendix I for details about the experiments)\footnote{All experimental procedures were performed in accordance with the Canadian Council on Animal Care guidelines and approved by the University of Western Ontario Animal Use Subcommittee (AUP: 2018–071D).}. We then implemented the short-term memory task where $1$ of $8$ items is to be held in memory (Fig. \ref{fig:short_term_memory}). As before, the cv-NN holds an item in short-term memory through the position of the phase coherent cluster. We then systematically injected dynamics from subsets of the cv-NN as a current into the biological neuron in separate trials, effectively using the biological cell in place of the $8$ decoding units used previously (Fig.~\ref{fig:biological_neurons}b). Inputs from the subset of the network corresponding to the remembered item sum constructively and cause the biological neuron to fire (black trace, Fig.~\ref{fig:biological_neurons}c), while inputs from outside the coherent phase cluster sum destructively and do not cause the neuron to fire (gray trace, Fig. \ref{fig:biological_neurons}c). Over several trials, the biological neuron repeatedly spiked successfully for the remembered item and not for other inputs (Fig.~\ref{fig:biological_neurons}d). These results are robust for different short-term memory items, different scaling factors to translate the cv-NN dynamics into a biological current (Supplement Fig.~S\revision{6}), and are consistent with a standard mathematical model of the neuron (Supplement Fig.~S\revision{7}). It is important to note that, in contrast to standard mathematical models of single neurons, which always fire a spike at a fixed threshold potential and instantaneously reset, biological neurons have variable thresholds that change dynamically in time and with different input \cite{platkiewicz2010threshold}. Even under these conditions, however, \revision{computations in the cv-NN can be successfully implemented by real biological cells. Taken together, these results open a new path to neuron-computer interfaces with exactly solvable dynamics.}

\section{Discussion}

\revision{In this paper, we introduce a cv-NN that can perform computations while also being exactly solvable. These results provide a comprehensive theoretical framework to understand computation with linear systems and nonlinear readout. These results unify previous experimental \cite{vandoorne2014experimental,lugnan2020photonic,vinckier2015high} and numerical observations \cite{laporte2018numerical} of computation in linear systems with nonlinear readout, in addition to providing a precise mathematical framework in which to design networks to perform computation. In this way, we leverage the exact mathematical description of spatiotemporal dynamics in the cv-NN to perform computations that can be precisely interpreted. This mathematical framework allows performing multiple types of tasks with the same network,} circumventing the need for computationally expensive training algorithms and hyperparameter tuning. This is, to the best of our knowledge, the first example of computation with spatiotemporal dynamics where such a closed-form expression \revision{-- precisely describing the whole dynamics-based computation, from the input to the output --} can be obtained.

Computing with linear network dynamics may at first seem counter-intuitive. Research in both neural networks and dynamics-based computation has \revision{largely} focused on nonlinear systems, both because real-world systems are in general nonlinear \cite{pathak2018model,tanaka2019recent,kia2020nonlinear,murali2022reconfigurable,choudhary2020physics} and because saturating nonlinearities can keep RNN activity within bounded intervals \cite{sompolinsky1988chaos,kadmon2015transition}. It is well known that RNNs can be trained to perform tasks similar to those we have studied here \cite{yang2019task,masoliver2022embedded,ehrlich2022geometry}. At the same time, however, it is in general difficult to tune RNNs to achieve desired task performance \cite{bengio1994learning} and that standard training algorithms, such as backpropagation through time (BPTT), are both difficult to train \cite{pascanu2013difficulty} and to interpret when successfully trained. In addition, previous work in dynamics-based computation with nonlinear systems generally required mapping the entire parameter space of the systems under consideration through exhaustive numerical simulation \cite{sinha1998dynamics,sinha1999computing,murali2009reliable}. As an alternative to standard RNNs, it is increasingly appreciated that nonlinear oscillator networks can be trained to perform sophisticated computations \cite{izhikevich2000computing,raychowdhury2018computing,heeger2019oscillatory,ricci2021kuranet,zanin2013computing,csaba2020coupled}. In this work, we have developed a mathematical framework that allows computation with linear oscillator dynamics and nonlinear readout, in a system which has the advantage that it can be exactly described in a closed-form expression. \revision{This provides a precise mathematical description of the dynamical trajectory the cv-NN uses to perform an individual computation, as it is generated through the combination of inputs, recurrent dynamics, time delays, and nonlinear readout.} In recent work, we have also developed a detailed analytical understanding of the connection between nonlinear oscillator networks and complex-valued linear systems \cite{muller2021algebraic,budzinski2022geometry,budzinski2023analytical}. This mathematical framework we have recently developed allowed us to understand how sophisticated spatiotemporal dynamics -- and, specifically, \revision{transient} chimera states -- can be found in networks of complex-valued oscillators with linear interactions. \revision{Importantly, our results are consistent with recent computational work on linear RNNs, which has shown that linear RNNs, interleaved with nonlinear multi-layer perceptrons, can show improved performance and efficiency compared to modern Transformer networks in certain key tasks \cite{orvieto2023resurrecting}.}

The computations we study here are possible in a linear system because we are considering the transient dynamics. Transient and spatiotemporal dynamics are highly useful for computation in artificial neural networks \cite{maass2002real,jaeger2004harnessing,buonomano2009state} and have been directly tied to computation in some biological neural systems \cite{mazor2005transient,la2019cortical,brinkman2022metastable}. Leveraging our analytical framework allowed us to construct this linear, complex-valued system where we can meaningfully compute with phase during the transient regime that extends for relatively long times. It is this technical advance that allows us to construct sophisticated computations with transient dynamics that can be precisely described by a mathematical framework. The results we introduce in this paper can thus provide a fundamentally new way of looking at computation. This network can perform non-trivial computations that can also be precisely specified by a closed-form mathematical expression. From this perspective, the results we report here can advance the level to which detailed analytical explanations of computation in neural networks may be possible.

Finally, the connection between working memory and dynamics-based computation identified in this work may provide insight into models of short-term memory in artificial and biological neural networks. Mathematical approaches to working memory in the executive regions of the brain are often based on static ``bump'' attractors \cite{chow2006existence,kilpatrick2013wandering}, where nodes within a bump of elevated activity are responsible for holding activity in working memory. Empirical results have shown, however, that more sophisticated dynamics, such as chaotic attractors, may play an important role in short-term memory processes \cite{wang202150}. The cv-NN developed here can be generalized to store short-term memories through arbitrary spatiotemporal patterns, beyond the partially synchronized chimera states that we utilized here and that bear resemblance to the bump attractor models of working memory. In this way, the particular dynamics-based short-term memory and computation studied in this work may possess many desirable features of in-memory computation \cite{sebastian2020memory} that has generated much interest as the next generation of computing hardware. The cv-NN introduced here may provide a guiding mathematical framework for implementing and training in-memory computing systems both in digital simulations and in physical hardware, with the advantage of the training being specified by precise mathematical expressions.

\begin{acknowledgments}
This work was supported by BrainsCAN at Western University through the Canada First Research Excellence Fund (CFREF), the NSF through a NeuroNex award (\#2015276), the Natural Sciences and Engineering Research Council of Canada (NSERC) grant R0370A01, Compute Ontario (computeontario.ca), Digital Research Alliance of Canada (alliancecan.ca), the Western Academy for Advanced Research, and NIH Grants U01-NS131914 and R01-EY028723. R.C.B gratefully acknowledges the Western Institute for Neuroscience Clinical Research Postdoctoral Fellowship.
\end{acknowledgments}

%\bibliography{references.bib}

\end{document}